\pdfoutput=1
\documentclass[10pt,twocolumn,letterpaper]{article}

\usepackage{cvpr}
\usepackage{times}
\usepackage{epsfig}
\usepackage{graphicx}
\usepackage{amsmath}
\usepackage{amssymb}

\usepackage[breaklinks=true,bookmarks=false]{hyperref}

\cvprfinalcopy 

\def\cvprPaperID{****} 

\begin{document}

\title{Naive-Deep Face Recognition: Touching the Limit of LFW Benchmark or Not?}

\author{
Erjin Zhou\\
Face++, Megvii Inc.\\
{\tt\small zej@megvii.com}
\and
Zhimin Cao\\
Face++, Megvii Inc.\\
{\tt\small czm@megvii.com}
 \and
 Qi Yin\\
Face++, Megvii Inc.\\
{\tt\small yq@megvii.com}
}

\maketitle

\begin{abstract}
Face recognition performance improves rapidly with the recent deep learning technique developing and underlying large training dataset accumulating.
In this paper, we report our observations on how big data impacts the recognition performance.
According to these observations, we build our Megvii Face Recognition System, which achieves 99.50\% accuracy on the LFW benchmark, outperforming the previous state-of-the-art.
Furthermore, we report the performance in a real-world security certification scenario. 
There still exists a clear gap between machine recognition and human performance.
We summarize our experiments and present three challenges lying ahead in recent face recognition.
And we indicate several possible solutions towards these challenges.
We hope our work will stimulate the community's discussion of the difference between research benchmark and real-world applications.

\end{abstract}

\section{INTRODUCTION}
The LFW benchmark \cite{LFWTech} is intended to test the recognition system's performance in unconstrained environment, which is considerably harder than many other constrained dataset (e.g., YaleB \cite{GeBeKr01} and MultiPIE \cite{gross2010multi}). 
It has become the de-facto standard regarding to face-recognition-in-the-wild performance evaluation in recent years. 
Extensive works have been done to push the accuracy limit on it \cite{cao2010face,yin2011associate,chen2012bayesian,berg2012tom,cao2013practical,chen2013blessing,sun2013hybrid,sun2014deep2,sun2014deep,taigman2014deepface,sun2014deeply,zhu2014recover,lu2014surpassing}.

Throughout the history of LFW benchmark, surprising improvements are obtained with recent deep learning techniques \cite{zhu2014recover,taigman2014deepface,sun2014deeply,sun2014deep2,sun2014deep}. 
The main framework of these systems are based on multi-class classification \cite{sun2014deep2,sun2014deep,taigman2014deepface,sun2014deeply}.
Meanwhile, many sophisticated methods are developed and applied to recognition systems 
(e.g., joint Bayesian in \cite{chen2012bayesian,cao2013practical,sun2014deep2,sun2014deep,sun2014deeply},
model ensemble in \cite{sun2014deep2,taigman2014deepface}, 
multi-stage feature in \cite{sun2014deep2,sun2014deep}, 
and joint identification and verification learning in \cite{sun2014deep2,sun2014deeply}).
Indeed, large amounts of outside labeled data are collected for learning deep networks.
Unfortunately, there is little work on investigate the relationship between big data and recognition performance.
This motivates us to explore how big data impacts the recognition performance.
 
Hence, we collect large amounts of labeled web data, and build a convolutional network framework. 
Two critical observations are obtained.
First, the data distribution and data size do influence the recognition performance.
Second, we observe that performance gain by many existing sophisticated methods decreases as total data size increases.

\begin{figure}[t]
\begin{center}
\includegraphics[width=1.05\linewidth]{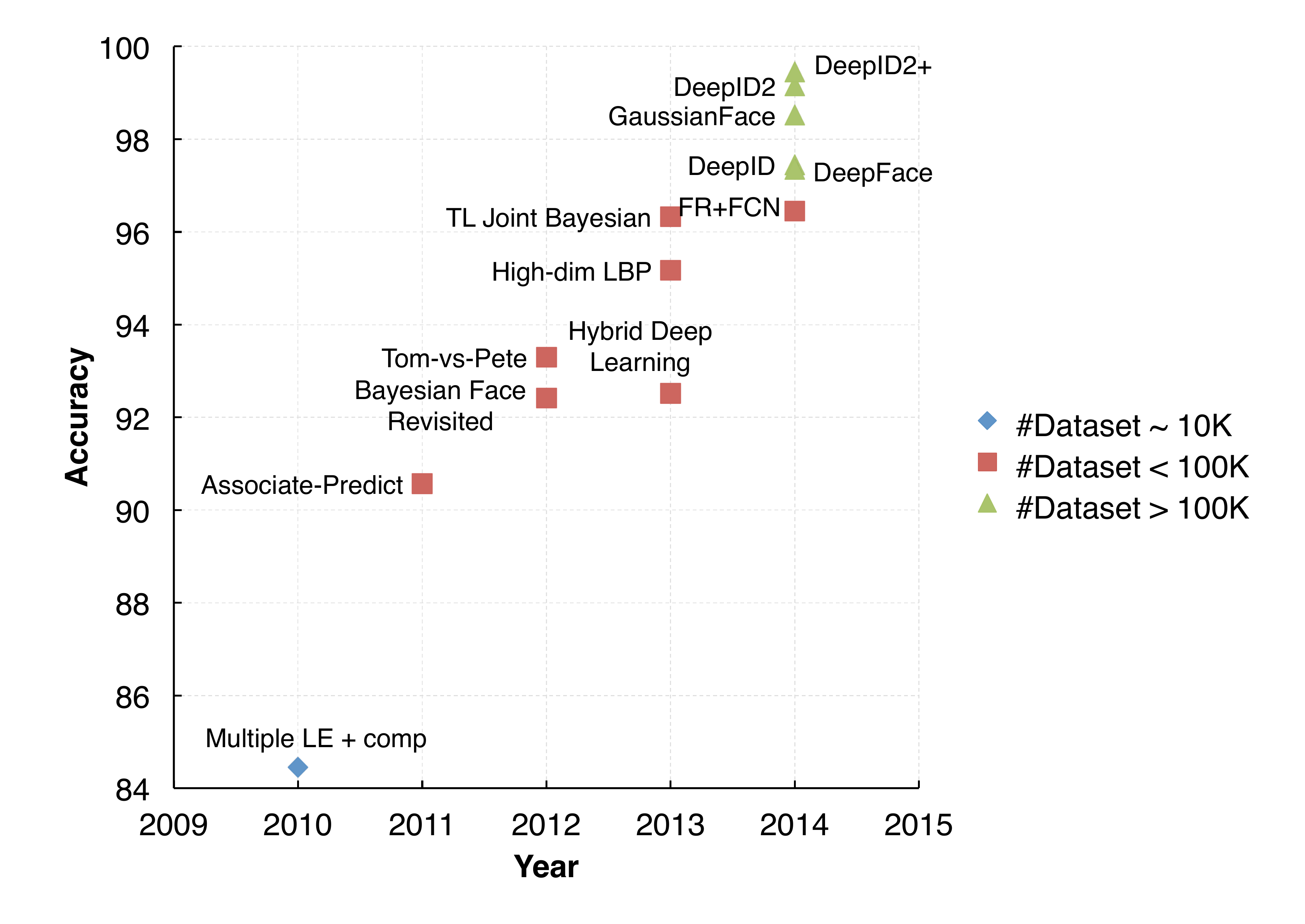}
\end{center}
\caption{{\bf A data perspective to the LFW history.}
Large amounts of web-collected data is coming up with the recent deep learning waves.
Extreme performance improvement is gained then.
How does big data impact face recognition?
}
\label{fig:lfw_history}
\end{figure}

According to our observations, we build our Megvii Face Recognition System by simple straightforward convolutional networks without any sophisticated tuning tricks or smart architecture designs.
Surprisingly, by utilizing a large web-collected labelled dataset, this naive deep learning system achieves state-of-the-art performance on the LFW.
We achieve the $99.50\%$ recognition accuracy, surpassing the human level.
Furthermore, we introduce a new benchmark, called Chinese ID (CHID) benchmark, to explore the recognition system's generalization. 
The CHID benchmark is intended to test the recognition system in a real security certificate environment which constrains on Chinese people and requires very low false positive rate.
Unfortunately, empirical results show that a generic method trained with web-collected data and high LFW performance doesn't imply an acceptable result on such an application-driven benchmark. 
When we keep the false positive rate in $10^{-5}$, the true positive rateis $66\%$, which does not meet our application's requirement.

By summarizing these experiments, we report three main challenges in face recognition: data bias, very low false positive criteria, and cross factors.  
Despite we achieve very high accuracy on the LFW benchmark, these problems still exist and will be amplified in many specific real-world applications.
Hence, from an industrial perspective, we discuss several ways to direct the future research.
Our central concern is around data: how to collect data and how to use data.
We hope these discussions will contribute to further study in face recognition.

\section{A DATA PERSPECTIVE TO FACE RECOGNITION}
An interesting view of the LFW benchmark history (see Fig.~\ref{fig:lfw_history}) displays that an implicitly data accumulation underlies the performance improvement.
The amount of data expanded 100 times from 2010 to 2014 (e.g., from about 10 thousand training samples in Multiple LE \cite{cao2010face} to 4 millions images in DeepFace \cite{taigman2014deepface}).
Especially, large amounts of web-collected data is coming up with the recent deep learning waves and huge performance improvement is gained then.

We are interested in this phenomenon. 
How does big data, especially the large amounts of web-collected data, impacts the recognition performance?

\section{MEGVII FACE RECOGNITION SYSTEM}

\subsection{Megvii Face Classification Database.} 
We collect and label a large amount of celebrities from Internet, referred to as the Megvii Face Classification (MFC) database. 
It has 5 million labeled faces with about 20,000 individuals. 
We delete all the person who appeared in the LFW manually.
Fig.~\ref{fig:data}~(a) shows the distribution of the MFC database, which is a very important characteristic of web-collected data we will describe later.

\subsection{Naive deep convolutional neural network.}
We develop a simple straightforward deep network architecture with multi-class classification on MFC database.
The network contains ten layers and the last layer is softmax layer which is set in training phase for supervised learning.
The hidden layer output before the softmax layer is taken as the feature of input image.
The final representation of the face is followed by a PCA model for feature reduction. 
We measure the similarity between two images through a simple L2 norm.

\section{CRITICAL OBSERVATIONS}
We have conducted a series experiments to explore data impacts on recognition performance.
We first investigate how do data size and data distribution influence the system performance.
Then we report our observations with many sophisticated techniques appeared in previous literatures, when they come up with large training dataset.
All of these experiments are set up with our ten layers CNN, applying to the whole face region.

\subsection{Pros and Cons of web-collected data}
\begin{figure}[t]
\begin{center}
\includegraphics[width=1.0\linewidth]{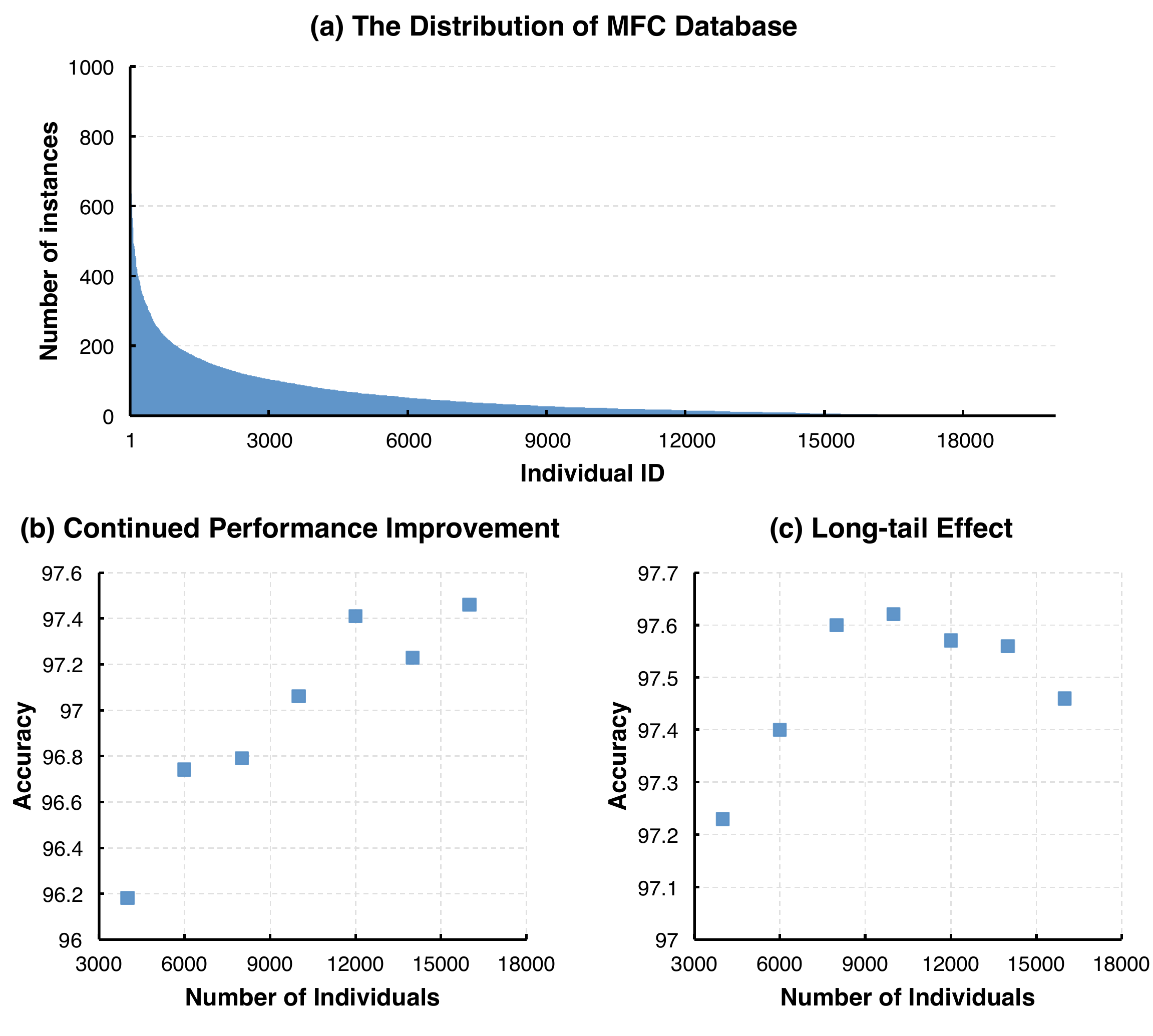}
\end{center}
\caption{{\bf Data talks.} 
(a) The distribution of the MFC database. All individuals are sorted by the number of instances. 
(b) Performance under different amounts of training data. 
The LFW accuracy rises linearly as data size increases.
Each sub-training set chooses individuals randomly from the MFC database. 
(c) Performance under different amounts of training data, meanwhile each sub-database chooses individuals with the largest number of instances. 
Long-tail effect emerges when number of individuals are greater than 10,000: keep increasing individuals with a few instances per person does not help to improve performance.
}
\label{fig:data}
\end{figure}

Web-collected data has typical long-tail characteristic: A few ``rich'' individuals have many instances, and a lot of individuals are ``poor'' with a few instances per person (see Fig.~\ref{fig:data}(a)).
In this section, we first explore how total data size influence the final recognition performance.
Then we discuss the long-tail effect in the recognition system.

{\bf Continued performance improvement.}
Large amounts of training data improve the system's performance considerably. 
We investigate this by training the same network with different number of individuals from 4,000 to 16,000.
The individuals are random sampled from the MFC database. 
Hence, each sub database keeps the original data distribution.
Fig.~\ref{fig:data}~(b) presents each system's performance on the LFW benchmark.
The performance improves linearly as the amounts of data accumulates.

{\bf Long tail effect.}
Long tail is a typical characteristic in the web-collected data and we want to know the impact to the system's performance.
We first sort all individuals by the number of instances, decreasingly.
Then we train the same network with different number of individuals from 4,000 to 16,000.
Fig.~\ref{fig:data}~(c) shows the performance of each systems in the LFW benchmark.
Long tail does influence to the performance.
The best performance occurs when we take the first 10,000 individuals with the most instances as the training dataset.
On the other words, adding the individuals with only a few instances do not help to improve the recognition performance.
Indeed, these individuals will further harm the system's performance.

\subsection{Traditional tricks fade as data increasing.}
We have explored many sophisticated methods appeared in previous literatures and observe that as training data increases, little gain is obtained by these methods in our experiments.
We have tried:\\
\noindent $\bullet$ {\bf Joint Bayesian}: modeling the face representation with independent Gaussian variables \cite{chen2012bayesian,cao2013practical,sun2014deep2,sun2014deep,sun2014deeply};\\
\noindent$\bullet$ {\bf Multi-stage features}: combining last several layers' outputs as the face representation  \cite{sun2014deep2,sun2014deep};\\
\noindent$\bullet$ {\bf Clustering}: labeling each individuals with the hierarchical structure and learning  with both coarse and fine labels \cite{yan2014hd};\\
\noindent$\bullet$ {\bf Joint identification and verification}: adding pairwise constrains on the hidden layer of multi-class classification framework \cite{sun2014deep2,sun2014deeply}.

All of these sophisticated methods will introduce extra hyper-parameters to the system, which makes it harder to train.
But when we apply these methods to the MFC database by trial and error, according to our experiments, little gain is obtain compared with the simple CNN architecture and PCA reduction.

\begin{figure}[t]
\begin{center}
\includegraphics[width=1.05\linewidth]{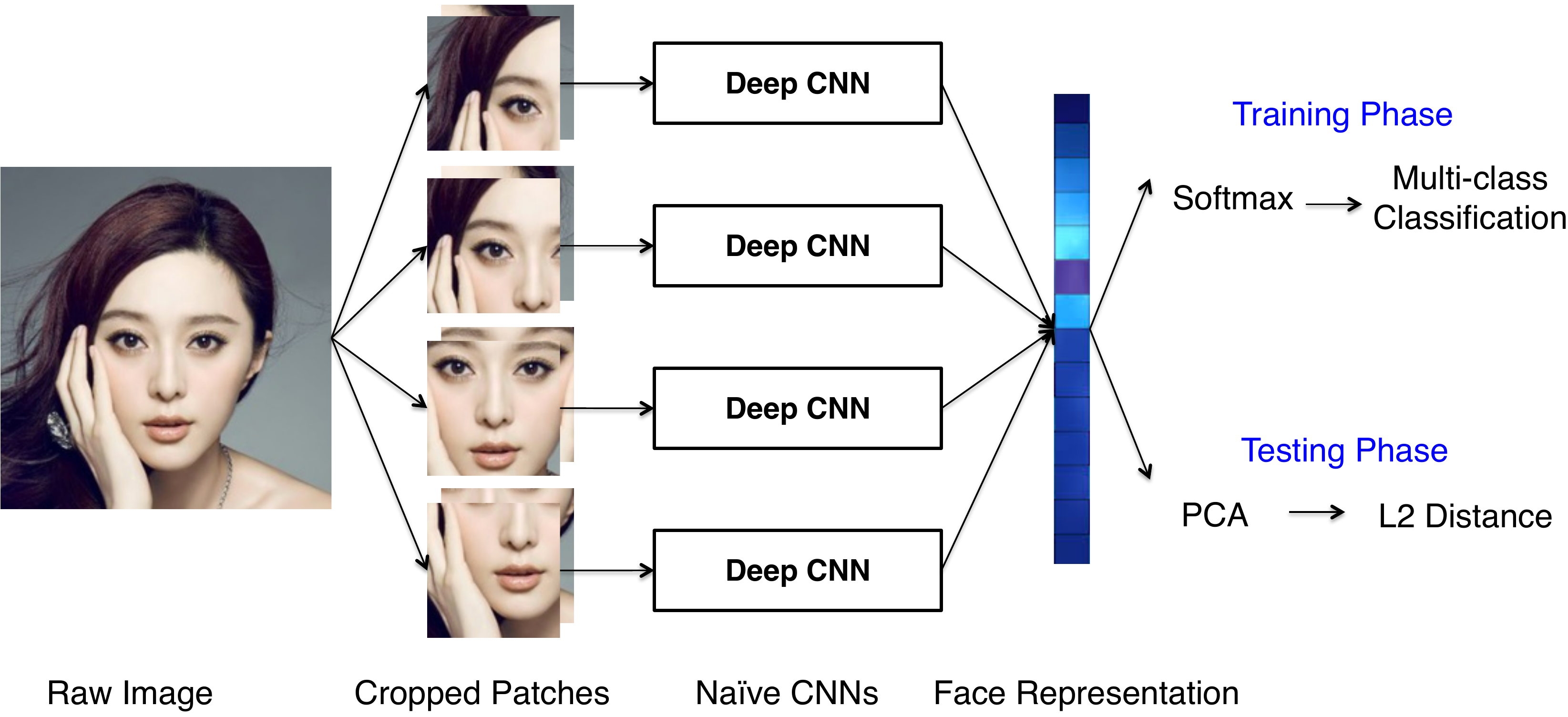}
\end{center}
\caption{{\bf Overview of Megvii Face Recognition System.}
We design a simple 10 layers deep convolutional neural network for recognition.
Four face regions are cropped for representation extraction.
We train our networks on the MFC database under the traditional multi-class classification framework.
In testing phase, a PCA model is applied for feature reduction, and a simple L2 norm is used for measuring the pair of testing faces.
}
\label{fig:system}
\end{figure}

\section{PERFORMANCE EVALUATION}
In this section, we evaluate our system to the LFW benchmark and a real-world security certification application.
Based on our previous observations, we train the whole system with 10,000 most ``rich'' individuals. 
We train the network on four face regions (i.e., centralized at eyebrow, eye center, nose tip, and mouth corner through the facial landmark detector).
Fig.~\ref{fig:system} presents an overview of the whole system.
The final representation of the face is the concatenation on four features and followed by PCA for feature reduction.

\subsection{Results on the LFW benchmark}
\begin{figure*}[t]
\begin{center}
\includegraphics[width=1.05\linewidth]{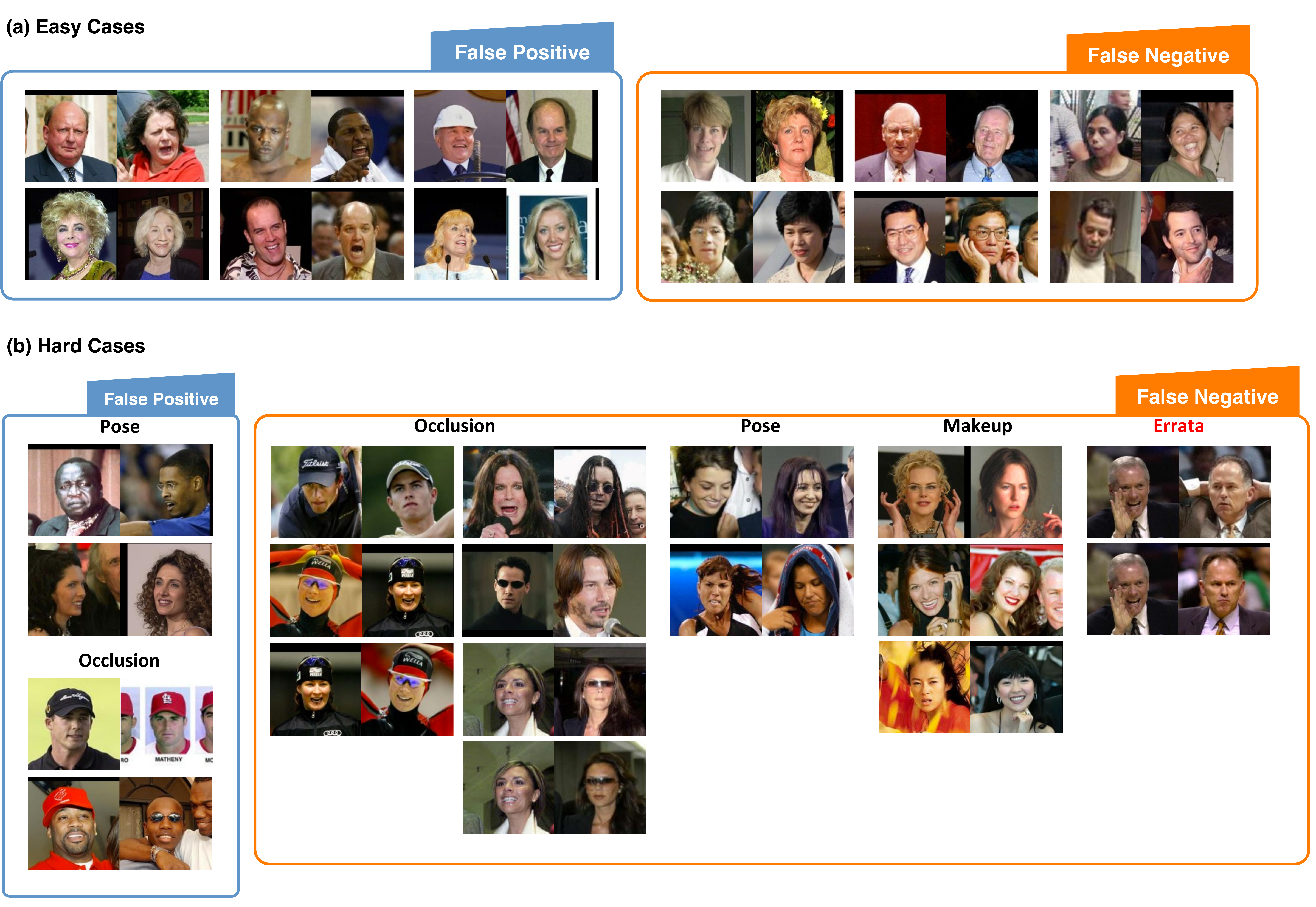}
\end{center}
\caption{{\bf 30 Failed Cases in the LFW benchmark.}
We present all the failed cases, and group them into two parts.
(a) shows the failed cases regarded as ``easy cases'', which we believe can be solved with a better training system under the existing framework.
(b) shows the ``hard cases''. These cases all present some special cross factors, such as occlusion, pose variation, or heavy make-up. 
Most of  them are even hard for human.
Hence, we believe that without any other priors, it is hard for computer to correct these cases. 
}
\label{fig:lfw_result}
\end{figure*}
We achieve $99.50\%$ accuracy on the LFW benchmark, which is the best result now and beyond human performance.
Fig.~\ref{fig:lfw_result} shows all failed cases in our system. 
Except for a few pairs (referred to as ``easy cases''), most cases are considerably hard to distinguish, even from a human.
These ``hard cases'' suffer from several different cross factors, such as large pose variation, heavy make-up, glass wearing, or other occlusions. 
We indicate that, without other priors (e.g., We have watched \textit{The Hours}, so we know that brown hair `` Virginia Woolf'' is Nicole Kidman), it's very hard to correct the most remain pairs.
Based on this, we think a reasonable upper limit of LFW is about $99.7\%$ if all the ``easy cases'' are solved. 

\subsection{Results on the real-world application}
\begin{figure}[t]
\begin{center}
\includegraphics[width=1.0\linewidth]{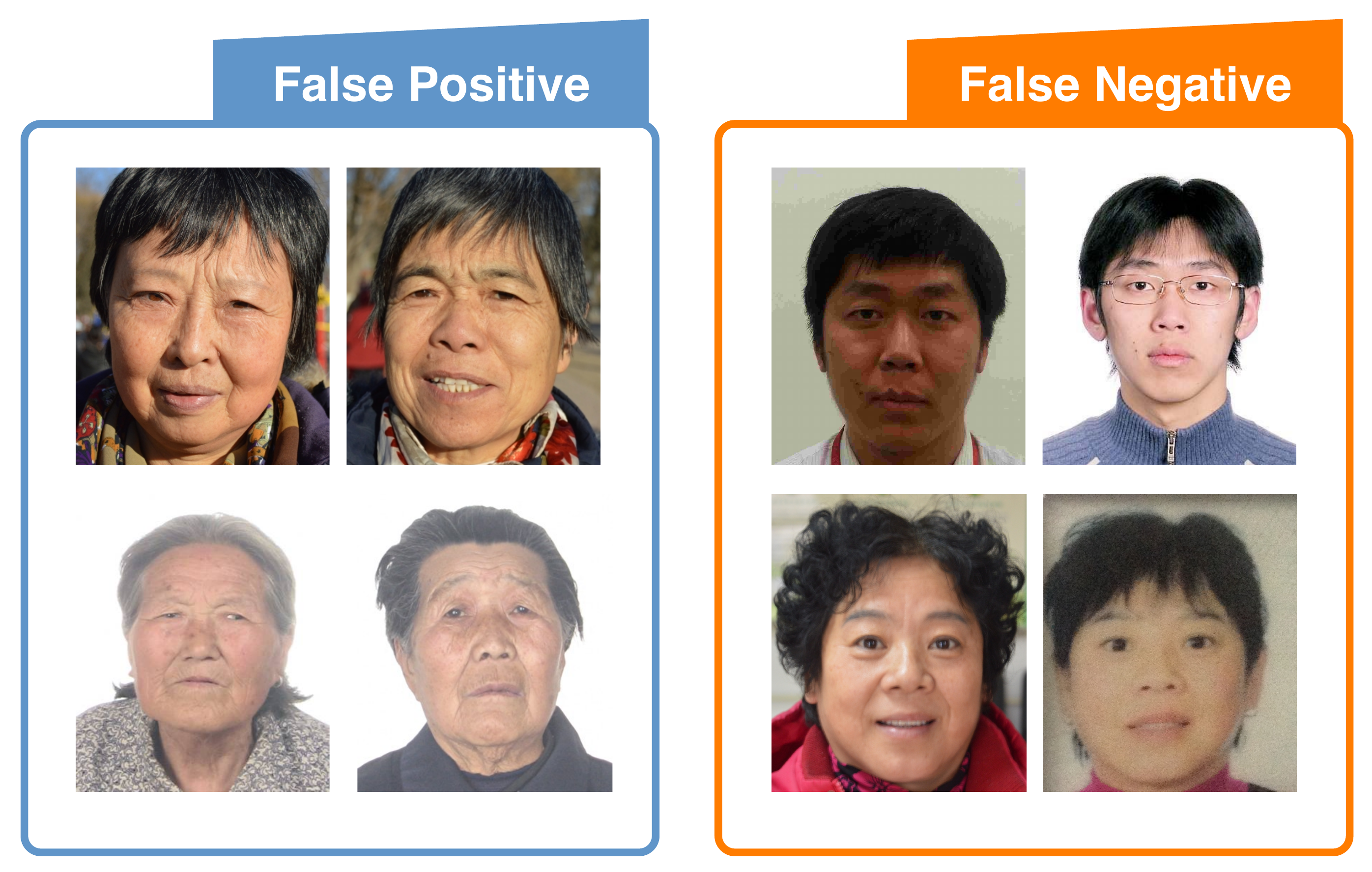}
\end{center}
\caption{{\bf Some Failed Cases in the CHID Benchmark.}
The recognition system suffers from the age variations in the CHID benchmark, 
including intra-variation (i.e., same person's faces captured in different age) and inter-variation (i.e., people with different ages).
Because little age variation is captured by the web-collected data, not surprisingly, the system cannot well handle this variation.
Indeed, we do human test on all these failed cases. Results show that $90\%$ failed cases can be solved by human.
There still exists a big gap between machine recognition and human level.
}
\label{fig:CHID_failed_cases}
\end{figure}
In order to investigate the recognition system's performance in real-world environment, we introduce a new benchmark, referred to as Chinese ID (CHID) benchmark. 
We collect the dataset offline and specialize on Chinese people. 
Different from the LFW benchmark, CHID benchmark is a domain-specific task to Chinese people. 
And we are interested in the true positive rate when we keep false positive in a very low rate (e.g., $F\!P=10^{-5}$). 
This benchmark is intended to mimic a real security certification environment and test recognition systems' performance.
When we apply our ``99.50\%'' recognition system to the CHID benchmark, the performance does not meet the real application's requirements.
The "beyond human" system does not really work as it seems.
When we keep the false positive rate in $10^{-5}$, the true positive rate is $66\%$.
Fig.~\ref{fig:CHID_failed_cases} shows some failed cases in $F\!P=10^{-5}$ criteria.
The age variation, including intra-variation (i.e., same person's faces captured in different age) and inter-variation (i.e., people with different ages), is a typical characteristic in the CHID benchmark.
Unsurprisingly,  the system suffers from this variation, because they are not captured in the web-collected MFC database.
We do human test on all of our failed cases.
After averaging 10 independent results, it shows $90\%$ cases can be solved by human, which means the machine recognition performance is still far from human level in this scenario.

\section{CHALLENGES LYING AHEAD}
Based on our evaluation on two benchmarks, here we summarize three main challenges to the face recognition.

{\bf Data bias.}
The distribution of web-collected data is extremely unbalanced. 
Our experiments show a amount of people with few instances per individual do not work in a simple multi-class classification framework.
On the other hand, we realize that large-scale web-collected data can only provide a starting point; it is a baseline for face recognition.
Most web-collected faces come from celebrities: smiling, make-up, young, and beautiful.
It is far from images captured in the daily life.
Despite the high accuracy in the LFW benchmark, its performance still hardly meets the requirements in real-world application.

{\bf Very low false positive rate.}
Real-world face recognition has much more diverse criteria than we treated in previous recognition benchmarks.
As we state before, in most security certification scenario, customers concern more about the true positive rate when false positive is kept in a very low rate.
Although we achieve very high accuracy in LFW benchmark, our system is still far from human performance in these real-world setting.

{\bf Cross factors.}
Throughout the failed case study on the LFW and CHID benchmark, pose, occlusion, and age variation are most common factors which influence the system's performance.
However, we still lack a sufficient investigation on these cross factors, and also lack a efficient method to handle them clearly and comprehensively.

\section{FUTURE WORKS}
Large amounts of web-collected data help us achieve the state-of-the-art result on the LFW benchmark, surpassing the human performance.
But this is just a new starting point of face recognition. 
The significance of this result is to show that face recognition is able to go out of laboratories and come into our daily life.
When we are facing the real-work application instead of a simple benchmark, there are still a lot of works we have to do.

Our experiments do emphasize that data is an important factor in the recognition system. 
And we present following issues as an industrial perspective to the expect of future research in face recognition.

On one hand, developing more smart and efficient methods mining domain-specific data is one of the important ways to improve performance.
For example,  video is one of data sources which can provide tremendous amounts of data with spontaneous weakly-labeled faces, but we have not explored completely and applied them to the large-scale face recognition yet.
On the other hand, data synthesize is another direction to generate more data.
For example, it is very hard to collect data with intra-person age variation manually. 
So a reliable age variation generator may help a lot.
3D face reconstruction is also a powerful tool to synthesize data, especially in modeling physical factors.

One of our observations is that the long-tail effect exists in the simple multi-class classification framework.
How to use long-tail web-collected data effectively is an interesting issue in the future.
Moreover, how to transfer a generic recognition system into a domain-specific application is still a open question.

This report provides our industrial view on face recognition, 
and we hope our experiments and observations will stimulate discussion in the community, both academic and industrial, and improve face recognition technique further.

{\scriptsize
\bibliographystyle{ieee}
\bibliography{face-recognition-tech-report}
}

\end{document}